%% file: main.tex
\documentclass[10pt,twocolumn,letterpaper]{article}

\usepackage[pagenumbers]{cvpr}
\input{preamble}

\usepackage{microtype}
\usepackage{booktabs}
\usepackage{amsmath,amssymb}
\usepackage{amsthm}
\newtheorem{proposition}{Proposition}
\theoremstyle{definition}

\definecolor{cvprblue}{rgb}{0.21,0.49,0.74}
\usepackage[pagebackref,breaklinks,colorlinks,allcolors=cvprblue]{hyperref}

\title{A Think Block Is Conditioning, Not a Plan}

\author{Jian Xu\textsuperscript{1} \qquad Delu Zeng\textsuperscript{2} \qquad Qibin Zhao\textsuperscript{1}\\
\textsuperscript{1}RIKEN AIP \qquad \textsuperscript{2}South China University of Technology\\
{\tt\small jian.xu@riken.jp}
}

\begin{document}
\maketitle

\begin{abstract}
Generative models increasingly write a chain of thought before drawing, and report large gains from
doing so. We ask what that text contributes, by substitution rather than ablation: holding the
generator, sampler and seeds fixed, we replace the think block with controlled alternatives and
measure the image. Across a unified model with a think mode, a reinforcement-learned reasoning
generator, and a planner driving a frozen diffusion model, the same picture emerges. Content-free
prose of matched length is worth nothing or less ($-0.05$ to $-0.15$). One sentence from an
unrelated 8B model, naming the subject and a few identifying details, matches the generator's own
plan in three to six times fewer conditioning tokens. The value sits where the referent is in doubt:
writing down a name is worth $+0.33$ when it is right and $+0.05$ when it is wrong, and where the
generator already succeeds the whole effect is $-0.07 \pm 0.07$. Cutting a plan at the text
encoder's window leaves 272 of 286 images byte-identical, and a plan rewritten to fit that window
ties a bare identification ($+0.003 \pm 0.044$), so a plan contains nothing the identification does
not already supply. Field-level interventions find only colour executed reliably, and a control
writing the same sentences into the user prompt shows that limit belongs to the generator rather
than to the planning channel. Two judges and two annotators replicate the central contrasts. A think
block earns its keep as a specification of what to draw, not as a plan that is carried out.
\end{abstract}

\section{Introduction}

A growing family of unified multimodal models writes a short planning passage before it renders an
image, and the reported gains are substantial: enabling an explicit think step raises BAGEL's WISE
score by $0.18$ \citep{deng2025bagel}, and related work builds reinforcement learning, interleaved
reasoning and test-time latent reasoning on the same premise \citep{jiang2025t2ir1,fang2025got,%
huang2025interleaving,mi2025milr}. The premise is that the passage is a \emph{plan}: a statement of
what the picture must contain, which the generator then carries out.

The premise has never been tested by intervention. Published evidence takes the form ``score with
reasoning minus score without'', which confounds the extra tokens, the knowledge the passage
recalls, and the composition it specifies. These are not idle alternatives: a parallel literature
improves the same benchmarks by rewriting the prompt with no reasoning claim at all
\citep{wu2025reprompt,wang2025promptenhancer}, and half a generator's reasoning tokens can be
removed with no loss \citep{gu2025shortcoti}. If the benefit were prompt expansion, the published
ablations would look exactly as they do.

We therefore ask the question by substitution rather than by ablation. Holding the generator, the
sampler, the resolution and the seeds fixed, we replace the contents of the think block with
controlled alternatives, each of which removes one candidate mechanism while preserving the others,
and we measure what happens to the image.

The generation case differs structurally from the language case, which is what makes it worth
testing rather than assuming. In a language model the chain of thought and the answer are produced
by one forward pass over one context, so the reasoning is causally upstream of the answer in a
direct sense. In a generative model the passage must pass through a conditioning channel, a prompt
string or a text embedding, into a renderer that was trained on captions. Whether a plan survives
that channel is an empirical question about the channel, not about reasoning, and the channel is
known to be lossy in specific ways \citep{zhang2024longclip,negt2i2026}.

One boundary belongs before any result. A substitution design measures what the contents of a think
block are worth to the renderer, not whether producing them required reasoning; a model that
resolves a riddle by several steps of inference and writes down the answer has reasoned. Nothing
here says reasoning is unnecessary. We claim that the renderer consumes a think block as a
specification rather than as a plan it carries out, and that scoring the model with reasoning
switched on and then off cannot tell those readings apart.

Our contributions are:

\begin{itemize}
\item an analysis (Section~\ref{sec:theory}) that separates what a think block can contribute into
occupancy, identification and elaboration, and identifies a pre-treatment variable, the model's own
identification, on which stratification turns the substitution arms into causal contrasts;
\item a decomposition of the published reasoning gain on WISE into those components, showing that
occupancy is zero, that the specification carries the effect, and that elaboration around a correct
specification is inert, replicated by two judges over every image and by two annotators on a paired
subset (Section~\ref{sec:wise});
\item the complementary measurement on GenEval, where nothing has to be recalled and the same
the task with the most room to improve, counting, is the one reasoning improves least
(Section~\ref{sec:geneval});
\item a field-level intervention study on three generators identifying which parts of a plan reach
the image, with a control that writes the same fields into the user prompt instead, separating a
limit of the planning channel from a limit of the generator (Section~\ref{sec:fields}).
\end{itemize}

\section{Background and related work}
\label{sec:background}

\paragraph{Reasoning-augmented generation} Unified models can emit text before pixels, and several
lines exploit it: a think mode that raises a world-knowledge score \citep{deng2025bagel},
reinforcement learning over semantic-level and token-level reasoning \citep{jiang2025t2ir1}, an
explicit chain carrying coordinates \citep{fang2025got}, interleaved text and image
\citep{huang2025interleaving}, and reasoning in latent space at test time \citep{mi2025milr}. They
share an evaluation protocol: a benchmark score with the reasoning enabled against the same model
with it disabled. Benchmarks built for reasoning-driven generation \citep{chen2025r2ibench,%
sun2025t2ireasonbench} report limited ability and, notably, that pipelines separating reasoning from
generation can beat joint ones.

\paragraph{Prompt rewriting} The confound is not hypothetical: rewriting the prompt improves the
same numbers with no claim about reasoning, whether the rewriter is trained against image-level
rewards \citep{wu2025reprompt} or against a fine-grained alignment reward
\citep{wang2025promptenhancer}. Conversely, rewarding brevity removes $54\%$ of an autoregressive
generator's reasoning tokens at no cost in quality \citep{gu2025shortcoti}, which is hard to
reconcile with those tokens being a plan that is executed. Our filler arm makes the comparison
direct, holding token count fixed and deleting content instead.

\paragraph{Faithfulness of reasoning} Whether a stated chain of thought is the computation that
produced the answer has a mature language-side methodology built on counterfactual edits
\citep{turpin2023language,lanham2023measuring}, read against a benign-perturbation baseline rather
than against the unperturbed condition alone \citep{comparedtowhat2026}. Visual \emph{understanding}
has had the same treatment \citep{liu2025faithfulvisual}. Visual \emph{generation} has not, and
because the reasoning must cross a conditioning channel into a separate renderer, the language-side
conclusions do not transfer.

\paragraph{The conditioning channel} What a text encoder transmits is itself studied: CLIP-based
conditioning truncates at 77 tokens and attends weakly to late content \citep{zhang2024longclip},
negation inverts \citep{negt2i2026}, and counting is a standing failure of compositional benchmarks
\citep{ghosh2023geneval}. These are reported as properties of prompts;
Section~\ref{sec:fields} shows they are equally properties of plans.

\section{What a think block can contribute}
\label{sec:theory}

\paragraph{Setup} Write $p$ for the user prompt, $t$ for the think block, and
$c = E(p,t)$ for the conditioning the renderer receives, where $E$ is the text encoder of a frozen
generator or the context of a unified model. An image is drawn as $x \sim G(\cdot \mid c)$ and
scored by a judge $s(x,p) \in [0, s_{\max}]$ that never sees $t$. Define the \emph{plan value}
\begin{equation}
V(t \mid p) \;=\; \mathbb{E}\!\left[s(x,p) \mid c = E(p,t)\right] - \mathbb{E}\!\left[s(x,p) \mid c = E(p,\varnothing)\right],
\label{eq:value}
\end{equation}
the quantity that published ablations estimate for $t = T(p)$, the block the model writes for itself.

A think block carries three things that \eqref{eq:value} does not separate: it \emph{occupies} the
channel with tokens, it \emph{identifies} a referent, and it \emph{elaborates}. Writing
$t = (\iota, \varepsilon)$ for the identification and elaboration parts, our arms hold two fixed and
vary the third: a content-free block of matched length is
$(\varnothing, \varepsilon_{\text{filler}})$, a bare noun phrase is $(\iota, \varnothing)$, a
correct reference description is $(\iota^\star, \varepsilon_{\min})$, and that description followed
by the model's own prose is $(\iota^\star, \varepsilon_{\text{self}})$. Two elementary facts say
where each component can be seen; neither is deep, and their use is that each turns a question about
reasoning into a measurement.

\begin{proposition}[Truncation nullity]
\label{prop:trunc}
Suppose the encoder factors through a prefix, $E(u) = E(u_{1:K})$ for every input string $u$. Let
$t, t'$ be think blocks whose concatenations with $p$ agree on their first $K$ tokens. Then
$G(\cdot \mid E(p,t)) = G(\cdot \mid E(p,t'))$, so $V(t \mid p) = V(t' \mid p)$, and every semantic
field stated beyond position $K$ has plan effect and intervention effect exactly zero.
\end{proposition}

\noindent The proof, and a corollary on what a null measurement does and does not establish, are in
Appendix~\ref{app:proofs}.

Proposition~\ref{prop:trunc} is immediate, and that is the point: it converts a claim about
reasoning into one about the encoder, falsifiable by moving a sentence rather than rewriting it.
Section~\ref{sec:fields} performs exactly that manipulation and recovers the predicted collapse for
a CLIP-conditioned generator and its absence for a long-context one.

\begin{proposition}[What the stratified design identifies]
\label{prop:ident}
Let $R$ be any variable computed from the prompt before the think block is chosen, so that $R$ does
not depend on which arm is applied. Because the arm is set by the experimenter and the prompt, the
sampler and the seed are held fixed across arms, for every arm $a$ and every value $r$ the contrast
\begin{equation}
\Delta(a \mid r) \;=\; \mathbb{E}\!\left[s \mid a, R = r\right] - \mathbb{E}\!\left[s \mid \varnothing, R = r\right]
\label{eq:strat}
\end{equation}
is the average causal effect, within the stratum $\{R = r\}$, of replacing the think block by $a$.
\end{proposition}

\noindent Appendix~\ref{app:proofs} gives the proof in potential-outcome notation and quantifies
what the unstratified contrast would have estimated instead.

We take $R = \mathbf{1}\{\hat y = y^\star\}$ for the model's own one-line answer to ``name what this
refers to'', produced once per prompt by a call shared across arms and therefore pre-treatment. The
distinction matters: scoring an arm on the prompts it happens to identify correctly against the
no-plan arm on \emph{all} prompts would confound the think block's content with the difficulty of
the prompts on which identification succeeds. Within strata that confound is gone, and since the
reference arm supplies the correct referent by construction, including where the model cannot name
it, the effect of supplying it is estimated directly.

These two facts leave one prediction, which we also test. If a think block contributes by resolving
which object to draw, its value should concentrate on prompts whose referent the unaided generator
fails to recover and vanish on prompts it already renders correctly; if instead it contributes a
composition the generator executes, no such concentration is expected, since a correctly identified
object can still be arranged better. That test stratifies on an outcome rather than a pre-treatment
variable, so Proposition~\ref{prop:ident} does not cover it and regression to the mean is live: we
measure the stratifying score on one seed and every arm on a held-out seed, and read filler within
the same strata as a bound on whatever artefact remains.

\section{Experimental design}
\label{sec:design}

\paragraph{Generators and arms} The decomposition is run on BAGEL-7B-MoT \citep{deng2025bagel}, a
unified model with a think mode, at $1024\times1024$ with its published sampler settings and fixed
seeds. We reproduce its own think-then-generate path and replace only the contents of the think
block, so the system prompt, user prompt, sampler and guidance contexts are identical across
conditions. Section~\ref{sec:arch} adds T2I-R1 \citep{jiang2025t2ir1} and a Qwen3-8B
\citep{yang2025qwen3} planner driving a frozen SDXL \citep{podell2024sdxl}.

Writing $T(p)$ for the block the model writes for itself, the arms are: \emph{no plan}; \emph{filler},
content-free photographic advice padded to the word count of $T(p)$; \emph{own reasoning}, $T(p)$;
\emph{own name}, the model's one-line answer to ``name what this refers to''; \emph{external name}
and \emph{external description}, a noun phrase and a thirty-word sentence from Qwen3-8B, which has
never seen the generator; \emph{reference description}, the benchmark's own explanation; and
\emph{reference + elaboration}, that explanation followed by the model's expansion with its
identification removed. Filler is the benign-perturbation control \citet{comparedtowhat2026} argue
for: token count fixed, content deleted.

\paragraph{Benchmarks and scoring} WISE \citep{niu2025wise} describes an object without naming it
and supplies a reference explanation; we use 215 prompts over all six categories, 3110 scored
images. GenEval \citep{ghosh2023geneval} states the composition explicitly; we use 335 prompts over
five of its six tasks, omitting attribute binding, 2040 images. Arms cover nested rather than
identical prompt sets, so every gain we report is a paired difference over the cells both arms
share, averaged within prompt before averaging across prompts, with the standard error clustered at
the prompt level. WISE is scored on its three-point rubric by Qwen3-VL-8B \citep{qwen3vl} shown the
image and the reference explanation and nothing else, never the think block or the arm. GenEval is
scored by reimplementing its rules on Grounding DINO \citep{liu2024groundingdino}. Both are
reimplementations, so absolute values are not comparable with published leaderboards; every
comparison is between arms inside one protocol. Full settings, prompt templates and judge
instructions are in the supplement.

\section{What the think block is worth}
\label{sec:wise}

\begin{figure*}[t]
\centering
\includegraphics[width=\textwidth]{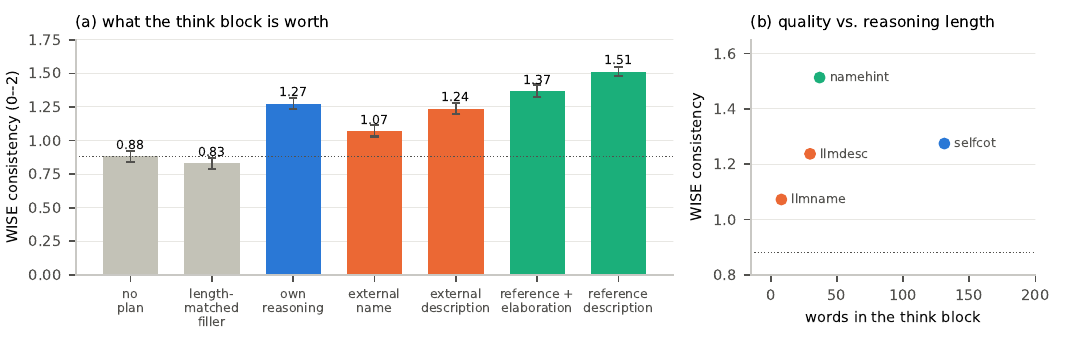}
\caption{(a) WISE consistency for each substitute think block, with standard errors; the dotted line
is the no-plan baseline. (b) The same scores against the length of the think block: quality is set by
what the block says, not by how long it is.}
\label{fig:wise}
\end{figure*}

\begin{figure*}[t]
\centering
\includegraphics[width=\textwidth]{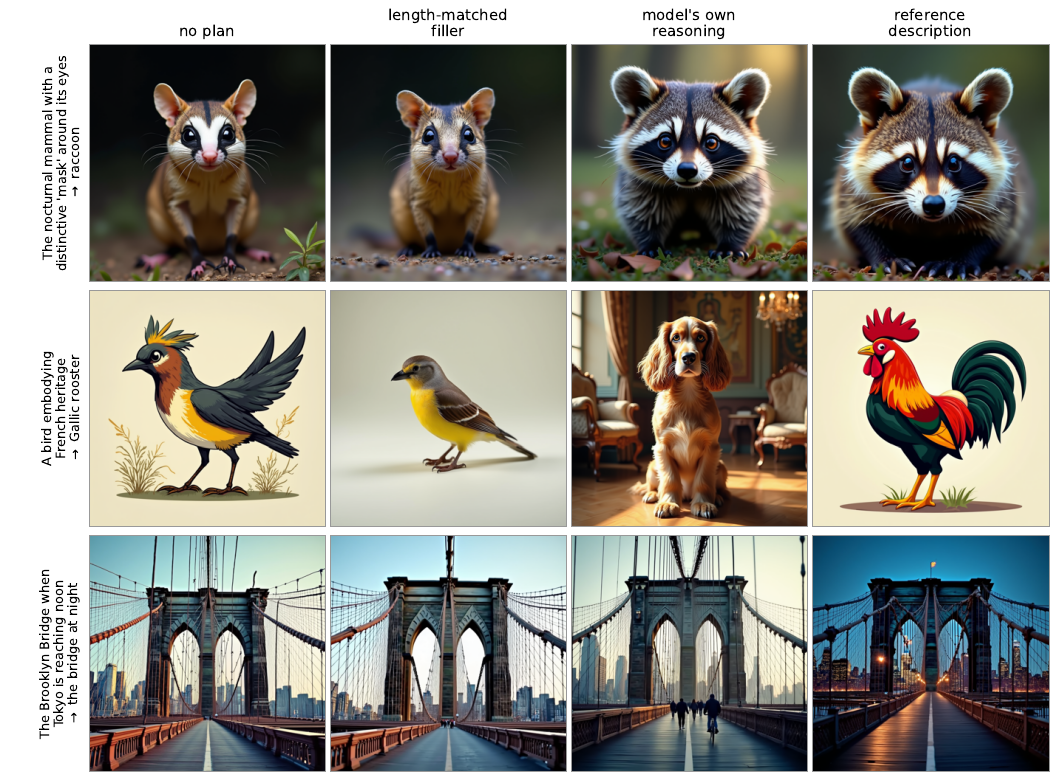}
\caption{Same generator, same prompt, same seed, four think blocks. Filler of the same length as the
model's own reasoning leaves the image where the no-plan arm left it. Reasoning helps when the model
names the referent correctly (row 1, raccoon) and not otherwise: it writes its way from ``a bird
embodying French heritage'' to a spaniel in a palace (row 2), and it keeps the daylight that the
time zone forbids at the Brooklyn Bridge (row 3). The reference description fixes both.}
\label{fig:qualwise}
\end{figure*}

Table~\ref{tab:wise} and Fig.~\ref{fig:wise} give the decomposition on WISE, and
Fig.~\ref{fig:qualwise} shows what the arms look like.

\begin{table}[t]
\centering
\caption{WISE consistency by arm (0--2 scale). Gains are paired differences against the no-plan arm.
Seeds of the same prompt are not independent, so each prompt's difference is averaged over its seeds
first and $n$ counts prompts, not images; the standard errors are therefore clustered at the prompt
level.}
\label{tab:wise}
\resizebox{\columnwidth}{!}{%
\begin{tabular}{lcccc}
\toprule
think block & words & $n$ & score & gain \\
\midrule
none & 0 & 215 & 0.87 & --- \\
length-matched filler & 139 & 215 & 0.82 & $-0.05 \pm 0.04$ \\
model's own reasoning & 131 & 215 & 1.27 & $+0.40 \pm 0.06$ \\
model's own name & 8 & 175 & 1.07 & $+0.20 \pm 0.05$ \\
external name (8B LLM) & 8 & 200 & 1.07 & $+0.18 \pm 0.04$ \\
external description (8B LLM) & 30 & 200 & 1.24 & $+0.34 \pm 0.05$ \\
reference description + elaboration & 132 & 150 & 1.37 & $+0.48 \pm 0.07$ \\
reference description & 37 & 215 & 1.51 & $+0.64 \pm 0.06$ \\
\bottomrule
\end{tabular}}

\end{table}

Three readings follow. First, \textbf{length is not the mechanism}: filler of the same length as the
model's own reasoning scores $0.83$ against the $0.88$ baseline, so occupying the channel with
plausible photographic language is worth nothing and if anything dilutes the prompt.

Second, \textbf{the effect is the specification, and it need not be internal}. One sentence from an
8B model that has never seen the generator scores $1.24$ against $1.27$ for the model's own 131-word
reasoning, a paired $-0.04 \pm 0.04$: a $4.4\times$ reduction in reasoning tokens costs nothing
measurable. A bare noun phrase recovers only half the gain ($1.07$), so what is useful is
identification plus a little visual description, not the noun alone.

Third, \textbf{a short correct identification matches a long chain of thought}. The benchmark's own
reference description, 37 words against the model's 131, reaches $1.51$. Under the judge this is
$+0.24 \pm 0.05$ above the model's own reasoning, but that gap is the one measurement our human
study does not reproduce: on a paired subset of 40 prompts two annotators put it at
$+0.03 \pm 0.11$ (Appendix~\ref{app:judge}), which excludes the judge's estimate at $2.2$ standard
errors. We therefore claim equality rather than superiority, which is the conservative reading and
is what the human data supports: \emph{the same quality at $28\%$ of the tokens}. Appending the
model's expansion to the correct description \emph{lowers} the judge's score by $0.13 \pm 0.05$, so
the prose written around an identification is at best inert.

\paragraph{The gain is identification, and it is causal within strata} Of the 175 prompts for which
the model's own one-line identification was judged, it names the referent correctly on $53\%$.
Because that answer is produced once per prompt by a call shared across arms, it is pre-treatment
and Proposition~\ref{prop:ident} applies: the contrasts in Table~\ref{tab:ident} are causal effects
of the think-block content within each stratum, not comparisons across populations of different
difficulty.

Writing down its own name is worth $+0.33 \pm 0.08$ when that name is right and $+0.05 \pm 0.05$
when it is wrong, so its value is entirely contingent on being correct, and a wrong name is close to
free rather than harmful. Supplying the correct referent to exactly the prompts the model
\emph{cannot} name is worth $+0.56 \pm 0.10$, nearly as much as on the prompts it can
($+0.67 \pm 0.09$); this is the contrast a conditioning account predicts and a reading on which the
model already had the knowledge does not. On those misnamed prompts its own reasoning recovers only
$+0.21 \pm 0.09$ of that $+0.56$.

\begin{table}[t]
\centering
\caption{WISE consistency within strata defined by the model's own identification, a variable fixed
before any think block is chosen. Parentheses give the paired gain over the no-plan arm in the same
stratum; standard errors of these gains, clustered at the prompt level, are $0.05$ to $0.10$.}
\label{tab:ident}
\resizebox{\columnwidth}{!}{%
\begin{tabular}{lcccc}
\toprule
stratum & none & own name & own reas. & reference \\
\midrule
right ($n{=}92$) & 0.94 & 1.27 ($+.33$) & 1.49 ($+.55$) & 1.61 ($+.67$) \\
wrong ($n{=}83$) & 0.80 & 0.86 ($+.05$) & 1.01 ($+.21$) & 1.36 ($+.56$) \\
\bottomrule
\end{tabular}}

\end{table}

\paragraph{The gain lives on prompts the generator gets wrong} Table~\ref{tab:strat} tests
the prediction of Section~\ref{sec:theory}. We stratify each prompt by its no-plan score on seed 1 and read every
arm on seed 2, so a prompt contributes one observation and cannot fall into two strata. Where the
unaided generator fails, the model's own reasoning is worth $+0.61 \pm 0.11$ and the reference
description $+0.99 \pm 0.11$. Where it already succeeds, they are worth $-0.07 \pm 0.07$ and
$+0.00 \pm 0.07$: nothing at all. Across prompts the correlation between the unaided score and the
gain is $-0.44$ for the reference description and $-0.13$ for filler.

Filler bounds whatever regression to the mean survives the held-out stratification, and the bound is
small: $+0.04 \pm 0.08$ in the failing stratum, against $+0.61$ and $+0.99$ for the content arms
there. In the upper two strata filler is negative, $-0.27 \pm 0.11$ and $-0.15 \pm 0.07$, which is
dilution of a prompt that was already sufficient and makes the content arms conservative rather than
inflated.

\begin{table}[t]
\centering
\caption{WISE consistency within strata. Each prompt is assigned to a stratum by its no-plan score
on seed 1 and every arm is read on seed 2, so each prompt appears once. Parentheses give the paired
gain with its standard error.}
\label{tab:strat}
\resizebox{\columnwidth}{!}{%
\begin{tabular}{lccccc}
\toprule
unaided & $n$ & none & filler & own & reference \\
\midrule
fails (0)    & 89 & 0.27 & 0.31 ($+.04$) & 0.88 ($+.61$) & 1.26 ($+.99$) \\
partial (1)  & 51 & 0.98 & 0.71 ($-.27$) & 1.39 ($+.41$) & 1.49 ($+.51$) \\
succeeds (2) & 60 & 1.88 & 1.73 ($-.15$) & 1.82 ($-.07$) & 1.88 ($+.00$) \\
\bottomrule
\end{tabular}}

\end{table}

\section{Does the decomposition hold across architectures?}
\label{sec:arch}

A think block reaches the renderer in different ways depending on how the system is built, and the
analysis of Section~\ref{sec:theory} says that difference should matter. We therefore repeat the
substitution design on two further generators that span the space.

T2I-R1 \citep{jiang2025t2ir1} is trained with reinforcement learning and reports a $19\%$
improvement on WISE from its semantic-level chain of thought; its own code builds the image prompt
as the user prompt followed by that chain, so the substitution applies unchanged. The third system
is not unified at all: Qwen3-8B \citep{yang2025qwen3} plans and writes the prompt for a frozen SDXL,
the decoupled arrangement \citet{chen2025r2ibench} report beats joint reasoning, and there a plan
enters only through a 77-token CLIP encoder.

\begin{table*}[t]
\centering
\caption{The same decomposition on three generators. Each entry is a paired difference against that
generator's own no-plan arm, with the mean length of the substituted block. The ordering of the arms
is identical in all three.}
\label{tab:arch}
\begin{tabular}{lcccccc}
\toprule
 & \multicolumn{2}{c}{BAGEL-7B-MoT} & \multicolumn{2}{c}{T2I-R1} & \multicolumn{2}{c}{Qwen3-8B + SDXL} \\
think block & gain & words & gain & words & gain & words \\
\midrule
no plan                 & ---              &   0 & ---              &   0 & ---              &   0 \\
length-matched filler   & $-0.05 \pm 0.04$ & 139 & $-0.10 \pm 0.04$ & 111 & $-0.15 \pm 0.04$ & 109 \\
the model's own plan    & $+0.40 \pm 0.06$ & 131 & $+0.27 \pm 0.05$ & 105 & $+0.18 \pm 0.04$ & 112 \\
external name (8B)      & $+0.18 \pm 0.04$ &   8 & $+0.16 \pm 0.04$ &  19 & $+0.14 \pm 0.04$ &  19 \\
external sentence (8B)  & $+0.34 \pm 0.05$ &  30 & $+0.41 \pm 0.05$ &  41 & $+0.30 \pm 0.05$ &  41 \\
reference description   & $+0.64 \pm 0.06$ &  37 & $+0.76 \pm 0.05$ &  48 & $+0.41 \pm 0.05$ &  48 \\
\bottomrule
\end{tabular}
\end{table*}

Table~\ref{tab:arch} gives the result, and three things hold in all three systems. Length-matched
filler is worth nothing and, on the frozen generator, less than nothing. A correct reference
description is the best block available, by $0.41$ to $0.76$ points. And one sentence from an 8B
model that has never seen the generator is worth at least as much as the generator's own plan,
$-0.04 \pm 0.04$ on the unified model, $+0.13 \pm 0.06$ on the reinforcement-learned one and
$+0.11 \pm 0.05$ on the planner pipeline.

The last two favour the external sentence under this judge, but not under a second one of a
different family scoring every image of both pipelines, where they are $+0.00 \pm 0.05$ and
$+0.05 \pm 0.05$: that judge scores the external sentence about $0.15$ lower while scoring the
generator's own plan almost identically. Everything else survives the change, filler negative and
the reference description best. We therefore claim equality and no more, \emph{a sentence from an
unrelated 8B model is worth at least as much as the generator's own plan, in three to six times
fewer conditioning tokens}, a claim about what the renderer needs rather than about total cost,
since an external planner substitutes one model's decoding for another's. The human replication on
the unified model agrees (Appendix~\ref{app:judge}).

The T2I-R1 column deserves comment, being the one system whose reasoning was explicitly trained. Its
published WISE gain is real here, $+0.27 \pm 0.05$, but a 41-word external sentence matches it and
the benchmark's own description reaches $+0.76 \pm 0.05$. Its reasoning prompt asks for a
restatement, instructing the model to include only what the user prompt states, and on a benchmark
whose prompts name nothing a faithful restatement cannot do the lookup the task needs.

\paragraph{Why the planner pipeline gains least from its own plan} The analysis predicts this and
the prediction is checkable without any new generation. Proposition~\ref{prop:trunc} says that text
beyond the encoder's window has no effect at all, and SDXL's encoder reads 77 CLIP tokens. The
planner's plans have a median length of 117 such tokens and $93\%$ of them exceed the window, so
most of what the planner writes cannot reach the renderer. What does reach it is determined by where
the identification sits, and the planner front-loads: it first names the referent by token 3 in half
of its plans, by token 19 in $95\%$ of them, and by token 37 in all of them. The surviving prefix is
therefore the identification and the discarded remainder is the elaboration, which is exactly why a
19-word name and a 112-word plan are worth the same on this architecture ($+0.14$ against $+0.18$)
while a 41-word sentence that fits the window entirely is worth more ($+0.30$). On the long-context
unified model, where nothing is discarded, the extra words do buy something, $+0.18$ for a name
against $+0.39$ for the full plan.

\subsection{Acting on the mechanism}
\label{sec:loop}

If the account above is right it is actionable, and it is falsifiable in two ways that cost nothing
but a rerun. Both are run on the planner pipeline, where the window is the binding constraint.

\paragraph{The window is real, not a figure of speech} Proposition~\ref{prop:trunc} says a plan and
the same plan cut at the encoder's window produce the same image. Cutting each plan at 75 CLIP
tokens, the paired difference is $+0.002 \pm 0.005$, and the reason is not statistical: $272$ of
$286$ regenerated images are byte-identical to the originals although the two conditioning strings
differ in every case, the truncated one shorter and lowercased by the tokenizer round trip. A
101-word plan and its 73-word prefix are the same input to this generator; whatever the planner
writes after its 77th token does not exist.

\paragraph{A plan written to fit does better} The planner wastes that room by default: its plans run
to a median of 117 CLIP tokens and $93\%$ exceed the window. Told how much room it has, in one added
clause and with nothing else changed, it writes 47 tokens instead and the gain rises from
$+0.181 \pm 0.037$ to $+0.279 \pm 0.040$, a paired improvement of $+0.098 \pm 0.035$. Less than half
the tokens, a better picture, and the only thing that changed is that the planner was told the size
of the channel it is writing into.

\paragraph{But the ceiling is identification, not packing} The same experiment bounds what this can
buy. A plan written to fit the window scores $+0.279 \pm 0.040$; a single sentence from the same
model naming the referent with a few identifying details scores $+0.295 \pm 0.040$, a paired
$+0.003 \pm 0.039$. Packing the channel with a full composition is worth exactly what naming the
subject is worth. This is the strongest form of our claim, because it could have failed: if a plan
held anything of value beyond the identification, compressing it to fit should have beaten a bare
identification. It does not, and the benchmark's own description still beats both
($-0.135 \pm 0.038$), so the headroom is identification accuracy, not plan engineering.

\section{When there is nothing to look up}
\label{sec:geneval}

If the value of a think block is the specification it supplies, it should shrink on a benchmark
whose prompts already supply one. GenEval prompts do exactly that: ``a photo of two dogs'', ``a photo of a purple potted plant''.
Over 2040 scored images the model's own reasoning is worth $+0.023 \pm 0.011$ (0.880 to 0.904),
filler is worth $-0.015 \pm 0.012$, and a think block that restates the required composition in
canonical form is worth $+0.023 \pm 0.010$. The largest single-tag effect is position
($0.848\to0.914$) and counting remains at $0.47$ in every arm.

A single cross-benchmark ratio is not meaningful here and we rest nothing on one: GenEval's no-plan
baseline is already $0.880$, leaving $0.120$ of its range, while WISE leaves $1.118$. Against the
headroom actually available, reasoning recovers $35.1\%$ on WISE and $19.4\%$ on GenEval, a factor
of $1.8$ rather than the $8.5$ a full-range normalisation would give. The ceiling-free comparison is
inside GenEval, where the tasks leave very unequal room: counting leaves the most, $0.553$, and
reasoning recovers $4.8\%$ of it, while position leaves $0.152$ and reasoning recovers $43.3\%$. The
task with the most to gain gains the least, matching the field measurements of
Section~\ref{sec:fields}. Attribute binding is absent from our sample, a limitation of the sample
and not of the benchmark; the per-task table is in the supplement.

\section{Which parts of a plan reach the image}
\label{sec:fields}

The GenEval result raises a sharper question: is compositional content simply not transmitted? We
test this directly with minimal interventions. We write 20 scenes whose user prompt is deliberately
under-specified (``a photo of a tidy kitchen counter, daylight'') and a plan that is the only source
of four semantic fields: the colour of a named object, a count, a left/right relation, and the
absence of an object. For each scene we generate from the reference plan, from the same plan with
exactly one field changed, from a meaning-preserving paraphrase, and from a plan that specifies none
of the four fields, three seeds each. A vision model answers one forced-choice question per field
from the image alone. We run the identical protocol on BAGEL and on SDXL, where the plan reaches the
generator only as prompt text through a 77-token CLIP window.

Table~\ref{tab:fields} gives the outcome on all three generators. Colour is transmitted by every one of them: stating it raises compliance by
$0.58$, $0.53$ and $0.45$, and editing it moves the image with probability $0.67$, $0.75$ and
$0.53$. Count is transmitted by none: the intervention effect is $0.10$, $0.12$ and $-0.02$, against
a stated change from two to five. The left/right relation is the one place where training the
reasoning appears to buy something, and even there not in the way one would want: T2I-R1 has the
largest intervention effect of the three ($0.43$ against $0.20$ and $-0.02$), yet its plan effect is
only $0.07$, so editing the field moves the image while stating it does not make the image more
likely to be right. That combination is sensitivity to wording without sensitivity to meaning, and
the paraphrase column confirms it: rewording the same plan costs T2I-R1 $0.37$ of colour compliance
and $0.25$ of count compliance, three times and twelve times what it costs the untrained unified
model. The system whose reasoning was optimised is the one most tightly bound to the exact words.
Two further effects are worth stating separately.

\begin{table}[t]
\centering
\caption{Field-level transmission on both architectures, 20 scenes $\times$ 3 seeds ($n=60$ per
cell). \emph{Plan} is compliance when the plan states the field minus compliance when it is silent;
\emph{edit} is the probability the image takes the new value after one field is changed, minus the
same probability under the unedited plan; \emph{para.} is the change in compliance under a
meaning-preserving paraphrase. Columns prefixed B are BAGEL-7B-MoT, R is T2I-R1 and S is SDXL.}
\label{tab:fields}
\resizebox{\columnwidth}{!}{%
\begin{tabular}{lccccccccc}
\toprule
field & B plan & B edit & B par. & R plan & R edit & R par. & S plan & S edit & S par. \\
\midrule
colour   & $+.58$ & $+.67$ & $-.10$ & $+.53$ & $+.75$ & $-.37$ & $+.45$ & $+.53$ & $-.17$ \\
count    & $+.17$ & $+.10$ & $-.02$ & $+.15$ & $+.12$ & $-.25$ & $+.10$ & $-.02$ & $-.10$ \\
spatial  & $+.02$ & $+.20$ & $-.18$ & $+.07$ & $+.43$ & $-.10$ & $-.07$ & $-.02$ & $-.02$ \\
absence  & $-.13$ & $+.37$ & $-.15$ & $\pm.00$ & $+.57$ & $-.08$ & $-.12$ & $+.27$ & $-.13$ \\
\bottomrule
\end{tabular}}

\end{table}

\paragraph{Negation is counterproductive} Writing ``there is no bottle anywhere in the scene'' into
the plan yields \emph{lower} compliance than saying nothing about bottles, on both architectures.
The reverse request works ($+0.37$ and $+0.27$ for making the object appear), so the field is
transmitted and it is the negation that is not, reproducing inside the reasoning channel the
polarity failure \citet{negt2i2026} document for direct prompts.

\paragraph{The channel has a measurable capacity} Proposition~\ref{prop:trunc} predicts that moving a
sentence past the encoder's window annihilates its effect exactly, and that the prediction can be
tested without changing a word. Moving the same four sentences to the end of the plan, behind 71
words of boilerplate, costs SDXL its only transmitted field: colour compliance falls from $0.67$ to
$0.20$, which is the no-plan prior of $0.22$. The same manipulation is harmless for the long-context
unified model ($0.80\to0.95$). The known truncation behaviour of CLIP-based conditioning
\citep{zhang2024longclip} is therefore not a nuisance of long prompts but the mechanism that decides
which parts of a plan exist for the generator at all. The same manipulation incidentally confirms the
negation result: truncating the ``no bottle'' sentence away returns compliance to the silent
baseline ($0.70\to0.81$).

\paragraph{Is this the plan's fault or the generator's?} A plan that is not executed could mean the
planning channel loses the instruction, or that the generator cannot render the field however it is
asked. The two are separable: we wrote the identical four clauses into the \emph{user prompt} and
gave the model the field-free plan, changing only where the text sits.

Table~\ref{tab:where} answers it in two parts. The planning channel does lose something: writing a
field into the prompt beats writing it into the plan on every field, by $+0.03$ to $+0.15$. But that
is second order. The first-order effect is that moving a field into the prompt does not rescue it.
Asked directly for exactly two objects the generator complies $0.43$ of the time against a prior of
$0.18$; asked directly for a left-right arrangement it reaches $0.70$ against $0.55$; asked directly
for an object to be \emph{absent} it does worse than if the object had never been mentioned, $0.73$
against $0.83$.

This is a difference of scale, not of kind, and we do not state it as a binary. Colour moves
compliance by $0.58$ to $0.73$ with $p<.001$ throughout. Count is not zero, $+0.17 \pm 0.08$ in the
plan and $+0.25 \pm 0.11$ in the prompt, the latter significant, with an intervention effect of
$+0.10 \pm 0.05$. The spatial relation has no detectable plan effect ($+0.02 \pm 0.10$) but an
intervention effect of $+0.20 \pm 0.09$. Some compositional information reaches the image, an order
of magnitude less of it than of an attribute. Two caveats bound this: 20 scenes, 3 seeds and 4
fields support a statement about these fields on these generators rather than a general claim about
compositional planning; and whatever its size, the shortfall is not attributable to the planning
channel, since the identical sentences in the most direct position available leave the ranking
unchanged.

\paragraph{Self-written plans commit to almost nothing checkable} The plans BAGEL writes unprompted
are stylistic expansions rather than specifications: across 20 scenes none states a left/right
relation, two contain a number and three mention an absence, which is why the interventions must
inject structured plans.

\section{Is the judge trustworthy?}
\label{sec:judge}

Every number above is a difference between arms measured by an automatic judge, so the judge is a
premise. Rescored by a judge of a different family, all 3110 WISE images agree within one point on $0.968$ of
cases, with no pair of arms changing order. Four annotators labelled 168 images blind to the
arm: agreement among them is itself moderate (Fleiss $\kappa = 0.350$), and against that ceiling the
judge is not the weak link, reaching $\kappa = 0.729$ against their majority, higher than any human
pair. The third check cost us a claim. Two annotators labelled 40 fresh prompts under all four
principal arms, computing a human gain exactly as an automatic one; filler, the model's own
reasoning and the reference description replicate, but the gap \emph{between} the last two does not,
$+0.38 \pm 0.11$ by judge against $+0.03 \pm 0.11$ by annotators, whose interval excludes the
judge's estimate. Hence equality throughout, here and for the external sentence in
Section~\ref{sec:arch}.

\section{Discussion and conclusion}

A think block is read by the renderer as a specification, not as a plan. Its value is carried by
which object it names and the few attributes it gives, not by its length: matched-length filler is
worth nothing, a sentence from an unrelated 8B model is worth as much as the generator's own, and
the effect disappears on prompts the generator was going to get right anyway. The elaboration is
inert for a reason we can show rather than infer, since on a frozen generator most of it never
reaches the model and a plan rewritten to fit what does ties a bare identification. What a plan
states beyond attributes is not executed, and not because a plan is the wrong vehicle, since the
same sentences fail from the user prompt too.

Three consequences follow. A gain from ``enabling reasoning'' is uninterpretable without a
length-matched control, and with one the brevity result of \citet{gu2025shortcoti} stops being
anomalous. Where the task is knowledge-bound the specification can come from anywhere, and the
headroom left, up to $+0.76$ on T2I-R1, is accuracy of the specification rather than depth of the
reasoning. Where it is composition-bound, neither a better plan nor a better prompt is the lever,
because the generator does not execute those fields from text at all.

\paragraph{What this does not say} Our arms measure what a think block is worth once it exists, not
how its content arose; a model that resolves a riddle by several steps of inference and then writes
the answer has reasoned. That question is at least approachable: asked for the referent directly,
Qwen3-8B is right on $0.493$ of the 215 prompts, and allowed 149 words of reasoning first it reaches
$0.535$, a paired $+0.042 \pm 0.033$ that fixes 29 answers and breaks 20. On this model and this
benchmark reasoning buys little identification accuracy, and we do not generalise: the interval
reaches $+0.11$, one model is not a class of models, and nothing here bears on prompts needing
genuine multi-step inference.

\paragraph{Limitations} Systems passing structured control rather than text, such as the
coordinate-carrying chains of \citet{fang2025got}, are untested, and our analysis predicts the
decomposition should \emph{not} hold for them, since such an interface bypasses the channel we
measure; latent-space reasoning \citep{mi2025milr} has no text block to substitute at all. Our
judges and GenEval rules are reimplementations, so no absolute value here belongs against a
published leaderboard. The field study is four fields and twenty scenes, and our GenEval sample
omits attribute binding. SDXL is not a recent architecture; its 77-token encoder is what makes the
channel measurable and also what bounds how far that half transfers. Finally, the only arm that
beats a generator's own plan everywhere is the benchmark's own description, an oracle.

\clearpage
\section*{LLM Usage Statement}

\paragraph{Models studied, as distinct from tools used} The generative models audited here
(BAGEL-7B-MoT, T2I-R1), the external planner (Qwen3-8B) and the judges (Qwen3-VL-8B-Instruct and,
for the robustness checks, Qwen2.5-VL-7B-Instruct) are the subject matter of the paper, not writing
aids. Their prompts, decoding settings and per-item outputs are released with the code, and
Section~\ref{sec:judge} quantifies how far the judges can be trusted. What follows concerns tools
used to produce the paper.

\paragraph{Which tools} Claude (Anthropic), run through a coding-agent interface with shell and file
access to the authors' compute cluster.

\paragraph{Which parts, and to what degree}
\begin{itemize}\itemsep2pt
\item \textbf{Research ideation: partial.} The topic and the substitution design are the authors'.
The assistant carried out the novelty search that redirected an earlier framing to the present one,
and proposed several of the controls, including the length-matched filler, the pre-treatment
stratification, the truncation test and the prompt-position control.
\item \textbf{Code: substantial.} All generation, substitution, scoring and analysis scripts were
written by the assistant under the authors' direction and are released unmodified.
\item \textbf{Experiments: substantial.} The assistant launched and monitored every run on the
authors' cluster, and performed the statistical analysis, including the prompt-level clustering that
corrected the reported standard errors.
\item \textbf{Figures: substantial.} All figures were produced by the assistant from the released
per-item outputs.
\item \textbf{Text: substantial.} The assistant drafted and revised the manuscript, including this
statement. The authors edited throughout and decided what the paper claims.
\item \textbf{Related work: assisted, then verified.} Candidate references were found by the
assistant through web search; every arXiv identifier, title, author list and venue was checked
against the source page before citation, and one citation was corrected after a name collision
between two unrelated papers with the same short title.
\item \textbf{Human annotation: none.} The annotators are people. No model produced, filtered or
adjudicated any human label.
\end{itemize}

\paragraph{Effect on the results} Three claims were withdrawn after the authors directed the
assistant to test claims the paper was making: the advantage of an external sentence over a
generator's own plan, which did not survive a second judge; a cross-benchmark ratio normalised by
the wrong quantity; and a binary reading of the field study that the per-field statistics do not
support. The authors verified every number against the released per-item outputs and take full
responsibility for the paper's contents, including any errors.

{
    \small
    \bibliographystyle{ieeenat_fullname}
    \bibliography{refs}
}

\clearpage
\appendix
\section{Proofs}
\label{app:proofs}

\subsection{Proposition~\ref{prop:trunc}}

Fix a prompt $p$ and write $u(t)$ for the string the generator is conditioned on when the think
block is $t$, so that the image is drawn as $x \sim G(\cdot \mid E(u(t)))$ for a deterministic
encoder $E$. Suppose $E$ factors through a prefix of length $K$, that is $E(u) = E(u_{1:K})$ for
every $u$, and let $t, t'$ be two think blocks with $u(t)_{1:K} = u(t')_{1:K}$.

Then $E(u(t)) = E(u(t)_{1:K}) = E(u(t')_{1:K}) = E(u(t'))$, so the two conditioning vectors are the
same element of the encoder's output space. The image law $G(\cdot \mid c)$ depends on its input
only through $c$, hence $G(\cdot \mid E(u(t))) = G(\cdot \mid E(u(t')))$ as probability measures.
Every functional of the image law therefore coincides under $t$ and $t'$. Three such functionals are
used in Section~\ref{sec:fields}. For a semantic field $f$ with stated value $v$, the compliance
$\pi_f(t) = \Pr[\,x \text{ exhibits } v \mid t\,]$; the plan effect $\pi_f(t) - \pi_f(t_\varnothing)$
against a plan $t_\varnothing$ that is silent about $f$; and the intervention effect
$\Pr[\,x \text{ exhibits } v' \mid t_{f \to v'}\,] - \Pr[\,x \text{ exhibits } v' \mid t\,]$ for an
edit of $f$ from $v$ to an alternative $v'$. If the sentence stating $f$ lies entirely beyond
position $K$, then $t$ and $t_{f \to v'}$ agree on the first $K$ tokens, so the plan effect and the
intervention effect for $f$ are both exactly zero. \qed

\paragraph{Corollary and its converse} A measured intervention effect that is bounded away from zero
is a certificate that the sentence stating that field lies inside the encoder's effective window: a
field outside it cannot move the image at all. The converse does not hold. A null effect is
consistent both with truncation and with the field being inside the window but not transmitted, and
no amount of rewriting separates the two, because rewriting changes the content whose effect is
being measured. This is why Section~\ref{sec:fields} tests by \emph{moving} the sentences rather
than by changing them: the content is held exactly fixed and only the position relative to $K$
varies, so a drop to the silent prior isolates the encoder as the cause. Table~\ref{tab:late}
reports that measurement.

\subsection{Proposition~\ref{prop:ident}}

The design is within-subject: every arm is applied to every prompt it covers, with the same seeds.
Index a cell by $i = (\text{prompt}, \text{seed})$ and let $x_i(a)$ be the image that cell produces
under arm $a$, with $\varnothing$ the no-plan arm. Because the prompt, the sampler, the resolution
and the random seed are held fixed and only the think block differs, $x_i(a)$ and $x_i(\varnothing)$
are both observed for the same $i$, and the individual effect
\begin{equation}
\tau_i(a) \;=\; s\!\left(x_i(a)\right) - s\!\left(x_i(\varnothing)\right)
\label{eq:tau}
\end{equation}
is observed directly rather than imputed. Let $R_i$ be any variable that is a function of the prompt
of cell $i$ alone. Conditioning on $\{R = r\}$ then selects the \emph{same} set of cells in both
terms of \eqref{eq:strat}, so
\begin{equation}
\mathbb{E}\!\left[s \mid a, R = r\right] - \mathbb{E}\!\left[s \mid \varnothing, R = r\right]
\;=\; \mathbb{E}\!\left[\tau(a) \mid R = r\right],
\end{equation}
the average effect of replacing the think block by $a$ within that stratum. \qed

\paragraph{Why $R$ must not depend on the arm} Suppose instead $R$ is read off the arm itself, for
example $R = \mathbf{1}\{\text{the think block in arm } a \text{ names the referent correctly}\}$.
Then $\{R = 1\}$ denotes one set of cells in the first term and the whole sample in a contrast
written against an unconditional baseline, and the difference decomposes as
\begin{equation}
\begin{split}
\underbrace{\mathbb{E}[s \mid a, R{=}1] - \mathbb{E}[s \mid \varnothing]}_{\text{what is reported}}
&= \underbrace{\mathbb{E}[\tau(a) \mid R{=}1]}_{\text{effect}} \\[2pt]
&\quad + \underbrace{\mathbb{E}[s \mid \varnothing, R{=}1] - \mathbb{E}[s \mid \varnothing]}_{\text{selection}} .
\end{split}
\label{eq:selection}
\end{equation}
The selection term is the amount by which prompts whose referent the model identifies are easier to
draw than the average prompt, and it is not zero in our data: the no-plan arm scores $0.940$ on the
prompts the model names correctly against $0.874$ over all prompts in the same subset, a selection
term of $+0.066$. An analysis that used the unstratified contrast would therefore have reported
$1.266 - 0.874 = +0.392$ for the effect of the model writing down its own name, where the
within-stratum effect is $1.266 - 0.940 = +0.326$. We report the latter throughout.

In our design $R$ is the correctness of the model's answer to ``in at most twelve words, name what
the following description refers to'', asked once per prompt in a separate call before any arm is
run and shared by every arm. It is therefore a function of the prompt alone, and
Proposition~\ref{prop:ident} applies.

\section{Experimental detail}
\label{app:detail}

\subsection{Generators and sampling}

BAGEL-7B-MoT is run through its own inference path with the published generation hyper-parameters:
$1024\times1024$, 50 denoising steps, text guidance scale $4.0$, image guidance scale $1.0$,
guidance interval $[0.4, 1.0]$, timestep shift $3.0$, global CFG renormalisation with minimum $0$.
Think text is produced greedily (no sampling) with a 300-token budget for full think blocks and 40
tokens for one-line identifications. The classifier-free guidance contexts are built exactly as in
the released code, so an injected think block enters the conditioned branch only, as the model's own
would. SDXL base 1.0 is run at $1024\times1024$ with 30 steps and guidance scale $5.0$; for SDXL a
plan is concatenated to the user prompt as additional prompt text, which is the only channel it has.
Seeds are fixed per cell and shared across arms, so two arms differ in the think block and in
nothing else.

\subsection{Think-block substitution}

For a prompt $p$ the model first writes its own block $T(p)$ along its ordinary think-then-generate
path. The arms then replace the contents of that block while keeping the surrounding system prompt
and the think delimiters identical:

\begin{itemize}
\item \emph{no plan}: think mode off, no block at all.
\item \emph{filler}: sentences drawn cyclically from a fixed pool of eight content-free photographic
remarks (``The lighting is soft and even, with no harsh shadows and no blown-out highlights'', and so
on), concatenated until the word count of $T(p)$ is reached. The pool mentions no object, colour,
count or spatial relation.
\item \emph{own reasoning}: $T(p)$ verbatim.
\item \emph{own name}: ``This refers to $\langle \hat y \rangle$'', where $\hat y$ is the model's own
answer to the identification question above.
\item \emph{external name} and \emph{external description}: the same template with $\hat y$ replaced
by, respectively, a noun phrase and a sentence of at most thirty words produced by Qwen3-8B, which
has no connection to the image generator.
\item \emph{reference description}: ``This refers to $\langle E \rangle$'', where $E$ is the
benchmark's own explanation field.
\item \emph{reference + elaboration}: the reference description followed by $T(p)$ with its leading
identification removed, so that the elaboration is the model's own but the identification is correct.
\end{itemize}

\subsection{Field-intervention scenes}

Each of the twenty scenes is a short user prompt that is deliberately silent about four fields,
together with a plan that states all four. For the kitchen windowsill scene the user prompt is ``A
photo of a kitchen windowsill'' and the reference plan $A$ reads

\begin{quote}\small
\textit{I will compose this picture as follows: the watering can is green; there are exactly two
small potted plants; the glass jar is on the left side of the watering can; there is no cat anywhere
in the scene. The lighting is soft and even, the camera looks straight at the subject, and the
composition is clean and uncluttered.}
\end{quote}

The arms are $A$; $B_f$ for each field $f$, identical to $A$ except that one field takes an
alternative value (green $\to$ pink, two $\to$ five, left $\to$ right, ``there is no cat'' $\to$ ``a
cat is clearly visible''); $C$, a meaning-preserving paraphrase of $A$ with every content word of the
four fields retained but the sentence forms changed; and $D$, which keeps the generic closing
sentences of $A$ and says nothing about the four fields. The late-position arm $E_{\text{late}}$
contains the identical four sentences, moved behind 71 words of the generic material, so that the
plan's content is unchanged and only its position relative to a text encoder's window differs.

\subsection{Judges}

No judge ever sees a think block, a plan, an arm label or a prompt's provenance.

\emph{WISE.} Qwen3-VL-8B-Instruct is shown the image and the benchmark's reference explanation and
asked to rate the match on the benchmark's three-point rubric, answering with a single digit: $0$
``does not depict it at all'', $1$ ``depicts it only partially or with clear errors'', $2$ ``clearly
and correctly depicts it''. Decoding is greedy with a six-token budget, and the first digit in
$\{0,1,2\}$ is taken as the score.

\emph{Fields.} The same model answers one forced-choice question per field from the image alone, for
example ``What colour is the watering can? Answer with a single colour word'', ``How many small
potted plants are visible? Answer with a single number'', ``Is the glass jar on the left side or the
right side of the watering can?'', ``Is a cat visible in the image?''. Answers are normalised to a
canonical token before comparison with the field's stated value.

\emph{Identification.} Qwen3-8B is shown the model's noun phrase and the benchmark's reference
explanation and asked ``Do they refer to the same thing? Answer with one word, yes or no'', greedily,
with thinking disabled. The judgements are released with the code. This criterion is stricter than
lexical overlap: it rejects, for instance, ``bat'' against a reference naming a raccoon, and
``durian'' against one naming a Buddha's Hand fruit, both of which a word-overlap rule accepts.

\emph{GenEval.} We reimplement the benchmark's decision rules on Grounding DINO base with box
threshold $0.35$ and text threshold $0.25$. Object presence and counts come from the detected boxes
for each required class; relative position compares box centres; colour is read by the
vision-language model from the detected crop and accepted on substring agreement in either
direction. This replaces the benchmark's own detector, so absolute values are not comparable with
published GenEval numbers.

\section{Is the judge a usable instrument?}
\label{app:judge}

Every number in this paper is a difference between arms measured by an automatic judge, so the
judge's reliability is a premise rather than a detail. We test it in three ways: a second judge of a
different family over the whole image set, a human ceiling from four independent annotators, and a
paired human replication of the contrasts the argument rests on.

\subsection{A second judge}

All 3110 WISE images were rescored with Qwen2.5-VL-7B-Instruct under the identical prompt. The two
judges agree exactly on $0.674$ of images and within one point on $0.968$, with a quadratic-weighted
$\kappa$ of $0.685$. The second judge is more lenient in level (mean $1.449$ against $1.150$), but
the shift is uniform: every arm's gain is attenuated by roughly a sixth and no pair of arms changes
order. Reading the decomposition off the second judge gives filler $-0.072 \pm 0.030$, the model's
own reasoning $+0.301 \pm 0.039$, an external name $+0.152 \pm 0.033$, an external description
$+0.280 \pm 0.038$ and the reference description $+0.530 \pm 0.039$. The two stratified analyses of
Section~\ref{sec:wise} also reproduce: under the second judge a wrong self-name is worth
$+0.05 \pm 0.04$, identical to the first judge's estimate, and in the stratum where the unaided
generator already succeeds the model's own reasoning is worth $+0.00$.

\subsection{A human ceiling}

Four annotators independently labelled a sample of 168 images, drawn seven per (arm, judge score)
cell so that the sample spans the rubric rather than its easy end. The annotator saw the image and
the benchmark's reference explanation and nothing else: no arm label, no think block, no automatic
score.

Agreement among the humans is itself moderate. Pairwise exact agreement averages $0.576$ and
quadratic-weighted $\kappa$ averages $0.531$, with Fleiss $\kappa = 0.350$ over all four; the four
differ widely in strictness, with mean labels from $0.82$ to $1.26$. Against that ceiling the judge
is not the weak link. Its $\kappa$ against each individual annotator is $0.532$, $0.547$, $0.565$
and $0.581$, which lies inside the human range and above the human average, and against the
four-way majority consensus it reaches exact agreement $0.679$ and $\kappa = 0.729$, higher than any
human pair. Its correlation with the mean of the four is $0.698$, while each annotator correlates
with the mean of the other three at $0.587$ to $0.657$. Replacing the judge with a fifth human would
not make the measurement more reliable.

Because the sample is stratified on the judge's own score and the arms have different score
distributions, levels must be reweighted by the inverse sampling probability of each (arm, score)
cell before they are compared; without that correction the per-arm biases are distorted, in one arm
by more than $0.38$. Reweighted, the judge sits $0.067$ below the average human, and the per-arm
biases are statistically indistinguishable from one another ($\chi^2 = 2.15$, $7$ d.f.,
$p = 0.95$, spread $0.084$). This matters because every reported number is a within-prompt
difference between arms, in which a bias common to all arms cancels exactly. The test is not
powerful, however: with 21 images per arm it would not detect an arm-specific bias below about
$0.45$, which is larger than several of the effects we report. That is what the next experiment is
for.

\subsection{A paired human replication}

A second annotation round drops the stratification and pairs instead. Forty prompts, none of them
used in the first round, were labelled under all four principal arms by two annotators, with the
arms of one prompt kept at least 30 positions apart in a shuffled sequence so that the comparison is
not visible as adjacent panels. The human gain over the no-plan arm is then computed exactly as the
automatic one is, as a within-prompt, within-annotator difference, so the annotator's strictness
cancels and no assumption about the judge's bias is needed.

The design carries its own check. The reference-description arm supplies the correct referent by
construction, so its paired gain has to come out positive if an annotator is rating whether the
image shows the stated subject at all; both annotations satisfy it, at $+0.475 \pm 0.143$ and
$+0.700 \pm 0.144$.

\begin{table}[t]
\centering
\caption{Round two. Paired gains over the no-plan arm on the same 40 prompts, by two human
annotators and by the judge. The first three rows replicate; the fourth does not.}
\label{tab:human2}
\resizebox{\columnwidth}{!}{%
\begin{tabular}{lccc}
\toprule
contrast & humans & judge & diff. \\
\midrule
filler $-$ no plan          & $+0.08 \pm 0.10$ & $-0.10 \pm 0.09$ & $1.3\sigma$ \\
own reasoning $-$ no plan   & $+0.56 \pm 0.14$ & $+0.33 \pm 0.15$ & $1.2\sigma$ \\
reference $-$ no plan       & $+0.59 \pm 0.12$ & $+0.70 \pm 0.15$ & $0.6\sigma$ \\
reference $-$ own reasoning & $+0.03 \pm 0.11$ & $+0.38 \pm 0.11$ & $2.2\sigma$ \\
\bottomrule
\end{tabular}}

\end{table}

Table~\ref{tab:human2} gives the result. The claims the paper rests on survive. Filler is worth
nothing to human annotators as well as to the judge, with a confidence interval that contains zero.
The model's own reasoning carries a large gain, which the humans in fact put higher than the judge
does. A 37-word reference description reaches the same level, $+0.59 \pm 0.12$ against
$+0.56 \pm 0.14$ for 131 words of the model's own prose.

The fourth row does not replicate. The judge scores the reference description $0.38$ above the
model's own reasoning, a $3.4\sigma$ effect under its own measurement, while the humans put that
difference at $+0.03 \pm 0.11$, and both annotators independently return near zero ($-0.05$ and
$+0.10$). The human interval excludes the judge's estimate, so this is a disagreement rather than a
shortage of power. We report the conservative reading throughout: the short correct identification
\emph{matches} the long chain of thought rather than beating it. We note also what this implies
about the previous subsection: an arm-specific judge bias of about $0.35$ exists and the homogeneity
test, whose resolution was $0.45$, could not see it. A bias test on a stratified sample is not a
substitute for a paired replication.

\subsection{The second judge across architectures}

The paired replication covers one generator, so the two other pipelines were rescored in full with
the second judge instead. Table~\ref{tab:j2arch} gives the result, and it is the reason
Section~\ref{sec:arch} claims equality rather than superiority.

\begin{table}[t]
\centering
\caption{Every arm of the two further pipelines, scored by both judges. The ordering is preserved
throughout, but the advantage of the external sentence over the generator's own plan is not.}
\label{tab:j2arch}
\resizebox{\columnwidth}{!}{%
\begin{tabular}{llcc}
\toprule
system & arm & judge 1 & judge 2 \\
\midrule
T2I-R1 & filler              & $-0.10 \pm 0.04$ & $-0.06 \pm 0.05$ \\
       & its own plan        & $+0.27 \pm 0.05$ & $+0.25 \pm 0.05$ \\
       & external sentence   & $+0.41 \pm 0.05$ & $+0.25 \pm 0.05$ \\
       & reference           & $+0.76 \pm 0.05$ & $+0.54 \pm 0.06$ \\
       & \emph{sentence $-$ own plan} & $+0.13 \pm 0.06$ & $+0.00 \pm 0.05$ \\
\midrule
planner & filler             & $-0.15 \pm 0.04$ & $-0.17 \pm 0.04$ \\
$+$SDXL & its own plan       & $+0.18 \pm 0.04$ & $+0.13 \pm 0.05$ \\
       & external sentence   & $+0.30 \pm 0.05$ & $+0.18 \pm 0.06$ \\
       & reference           & $+0.41 \pm 0.05$ & $+0.35 \pm 0.05$ \\
       & \emph{sentence $-$ own plan} & $+0.11 \pm 0.05$ & $+0.05 \pm 0.05$ \\
\bottomrule
\end{tabular}}
\end{table}

Filler stays negative, the generator's own plan keeps its gain, and the reference description stays
far ahead. What does not survive is the contrast we would most have liked to report. The second
judge scores the external sentence about $0.15$ lower than the first does while scoring the
generator's own plan almost identically, so the margin between them collapses to
$+0.00 \pm 0.05$ and $+0.05 \pm 0.05$. This is the same kind of arm-specific judge bias the paired
human replication found on the unified model, of about the same size, and it lands on a different
arm. The lesson we draw is procedural: a between-arm difference of a couple of tenths, measured by
one judge, should not be reported as a finding. The claims in this paper are the ones that two
judges and, where we have them, two annotators agree on.

\begin{figure*}[t]
\centering
\includegraphics[width=\textwidth]{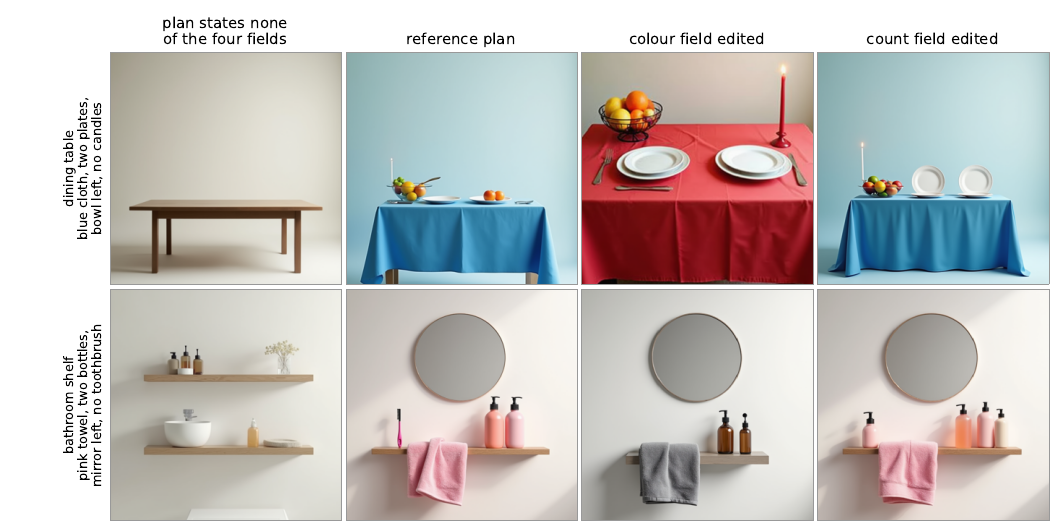}
\caption{Minimal edits of a single plan field, same generator and seed. Changing the colour word
changes the image; changing the count from two to five does not. In both reference plans the one
object the plan forbids is present anyway: a lit candle on the table that is to have no candles, a
toothbrush on the shelf that is to have none.}
\label{fig:qualfields}
\end{figure*}

\begin{table}[t]
\centering
\caption{The same four sentences in the plan (A) or in the user prompt (P), against the silent
control (D). Compliance on the unified model, 20 scenes $\times$ 3 seeds; differences are paired on
scene, with scene-clustered standard errors and two-sided $p$ values over the 20 scenes.}
\label{tab:where}
\resizebox{\columnwidth}{!}{%
\begin{tabular}{lccc}
\toprule
field & D silent & A in the plan & P in the prompt \\
\midrule
colour   & 0.22 & 0.80 ($+.58 \pm .08$, $p{<}.001$) & 0.95 ($+.73 \pm .09$, $p{<}.001$) \\
count    & 0.18 & 0.35 ($+.17 \pm .08$, $p{=}.06$)  & 0.43 ($+.25 \pm .11$, $p{=}.04$) \\
spatial  & 0.55 & 0.57 ($+.02 \pm .10$, $p{=}.87$)  & 0.70 ($+.15 \pm .11$, $p{=}.19$) \\
absence  & 0.83 & 0.70 ($-.13 \pm .07$, $p{=}.09$)  & 0.73 ($-.10 \pm .11$, $p{=}.38$) \\
\bottomrule
\end{tabular}}
\end{table}

\begin{figure*}[t]
\centering
\includegraphics[width=\textwidth]{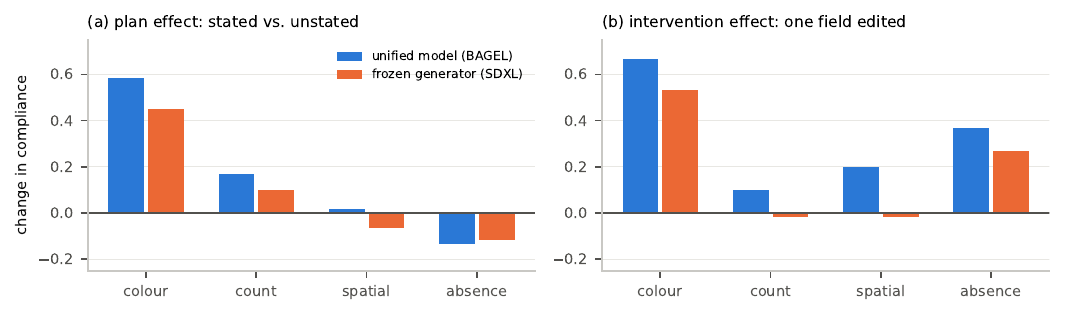}
\caption{Field-level transmission, 20 scenes $\times$ 3 seeds ($n=60$ per cell). (a) Compliance when
the plan states the field minus compliance when it says nothing about it. (b) Probability that the
image shows the new value after exactly one field of the plan is edited, minus the same probability
under the unedited plan.}
\label{fig:fields}
\end{figure*}

\section{Full breakdowns}
\label{app:tables}

\begin{table*}[t]
\centering
\caption{WISE consistency by category and arm. Each cell is the arm's mean on that category with, in
parentheses, its paired gain over the no-plan arm on the cells the two arms share. $n$ counts
no-plan images in the category.}
\label{tab:cats}
\resizebox{\textwidth}{!}{%
\begin{tabular}{lccccccccc}
\toprule
category & $n$ & none & filler & own reas. & own name & LLM name & LLM descr. & ref.+elab. & reference \\
\midrule
Biology            &  50 & 0.72 & 0.68 ($-.04$) & 1.14 ($+.42$) & 1.02 ($+.24$) & 0.98 ($+.25$) & 1.08 ($+.35$) & 1.31 ($+.47$) & 1.48 ($+.76$) \\
Chemistry          &  51 & 0.49 & 0.39 ($-.10$) & 0.80 ($+.31$) & 0.48 ($-.02$) & 0.50 ($-.02$) & 0.77 ($+.25$) & 1.14 ($+.61$) & 1.27 ($+.78$) \\
Cultural knowledge & 166 & 0.95 & 0.93 ($-.02$) & 1.51 ($+.56$) & 1.26 ($+.32$) & 1.28 ($+.31$) & 1.46 ($+.49$) & 1.43 ($+.50$) & 1.60 ($+.64$) \\
Physical knowledge &  49 & 0.98 & 0.73 ($-.24$) & 1.14 ($+.16$) & 1.02 ($+.10$) & 1.06 ($+.10$) & 1.10 ($+.15$) & 1.53 ($+.53$) & 1.53 ($+.55$) \\
Space              &  50 & 1.10 & 1.14 ($+.04$) & 1.36 ($+.26$) & 1.21 ($+.12$) & 1.10 ($-.02$) & 1.33 ($+.21$) & 1.64 ($+.47$) & 1.48 ($+.38$) \\
Time               &  49 & 0.90 & 0.88 ($-.02$) & 1.14 ($+.24$) & 1.00 ($+.14$) & 1.02 ($+.12$) & 1.15 ($+.25$) & 1.06 ($+.28$) & 1.53 ($+.63$) \\
\midrule
all                & 415 & 0.88 & 0.83 ($-.05$) & 1.27 ($+.39$) & 1.07 ($+.20$) & 1.07 ($+.18$) & 1.24 ($+.34$) & 1.37 ($+.48$) & 1.51 ($+.63$) \\
\bottomrule
\end{tabular}}
\end{table*}

Table~\ref{tab:cats} gives the category breakdown, and the ordering follows the specification
account. Chemistry is the category the model identifies worst, and there its own name is worth
nothing at all ($-0.02$) while the benchmark's description is worth $+0.78$, the largest gain in the
table: the channel is open, and what is missing is the referent. Cultural knowledge is the category
it identifies best, and there its own name already recovers half of what the reference achieves
($+0.32$ against $+0.64$). Filler is at or below zero in five of six categories.

\begin{table}[t]
\centering
\caption{GenEval accuracy by task. All arms cover an identical set of 510 cells over 335 prompts.}
\label{tab:tags}
\resizebox{\columnwidth}{!}{%
\begin{tabular}{lcccc}
\toprule
task & none & own reas. & filler & restate \\
\midrule
single object & 0.950 & 0.964 & 0.971 & 0.971 \\
two objects   & 0.904 & 0.924 & 0.866 & 0.927 \\
counting      & 0.447 & 0.474 & 0.461 & 0.474 \\
colours       & 0.931 & 0.941 & 0.915 & 0.952 \\
position      & 0.848 & 0.914 & 0.798 & 0.874 \\
\midrule
all           & 0.880 & 0.904 & 0.865 & 0.903 \\
\bottomrule
\end{tabular}}

\end{table}

\begin{table}[t]
\centering
\caption{The late-position arm. The same four sentences are moved behind 71 words of generic
material, with their content unchanged. For the frozen generator every field returns to the value it
takes when the plan is silent; for the long-context unified model none of them does. Columns
prefixed B are BAGEL-7B-MoT and S are SDXL.}
\label{tab:late}
\resizebox{\columnwidth}{!}{%
\begin{tabular}{lcccccc}
\toprule
field & B early & B late & B silent & S early & S late & S silent \\
\midrule
colour   & 0.80 & 0.95 & 0.22 & 0.67 & 0.20 & 0.22 \\
count    & 0.35 & 0.50 & 0.18 & 0.28 & 0.18 & 0.18 \\
spatial  & 0.57 & 0.50 & 0.55 & 0.45 & 0.48 & 0.52 \\
absence  & 0.70 & 0.65 & 0.83 & 0.70 & 0.82 & 0.82 \\
\bottomrule
\end{tabular}}

\end{table}

Table~\ref{tab:late} is the measurement that Proposition~\ref{prop:trunc} calls for. Moving the
sentences past the 77-token window of SDXL's text encoder returns all four fields to within $0.04$ of
the value they take when the plan never mentions them, which is what a plan that the encoder cannot
see must produce. The same move leaves the unified model's compliance unchanged or better. Since the
sentences are identical in both conditions, the difference between the architectures is a property
of the conditioning interface and not of the plan.

\end{document}

%% file: preamble.tex
\usepackage{caption} 

